\documentclass[letterpaper, 10 pt, conference]{ieeeconf}  

\IEEEoverridecommandlockouts                              

\usepackage{graphicx}
\usepackage{subcaption}
\usepackage[linesnumbered,ruled,vlined]{algorithm2e}
\usepackage{hyperref}
\usepackage{amsmath}
\usepackage{amssymb}
\DeclareMathOperator*{\argmax}{arg\,max}

\usepackage{biblatex}

\usepackage{booktabs}
\usepackage{multirow}
\usepackage{multicol}
\usepackage{makecell}
\usepackage{anyfontsize}

\title{\LARGE \bf
SOE: Sample-Efficient Robot Policy \underline{S}elf-Improvement \\via \underline{O}n-Manifold \underline{E}xploration\vspace{-4mm}}

\author{Yang Jin$^{1}$, Jun Lv$^{1,3}$, Han Xue$^{1}$, Wendi Chen$^{1,2}$, Chuan Wen$^{1\dagger}$, Cewu Lu$^{1,2,3\dagger}$\\
$^{1}$Shanghai Jiao Tong University $^{2}$Shanghai Innovation Institute $^{3}$Noematrix Ltd. $^\dagger$Equal Advising\\
\href{https://ericjin2002.github.io/SOE}{\texttt{ericjin2002.github.io/SOE}}\vspace{-4mm}
}

\begin{document}

\maketitle
\thispagestyle{empty}
\pagestyle{empty}

\begin{abstract}

Intelligent agents progress by continually refining their capabilities through actively exploring environments. Yet robot policies often lack sufficient exploration capability due to action mode collapse. Existing methods that encourage exploration typically rely on random perturbations, which are unsafe and induce unstable, erratic behaviors, thereby limiting their effectiveness. We propose Self-Improvement via On-Manifold Exploration (SOE), a framework that enhances policy exploration and improvement in robotic manipulation. SOE learns a compact latent representation of task-relevant factors and constrains exploration to the manifold of valid actions, ensuring safety, diversity, and effectiveness. It can be seamlessly integrated with arbitrary policy models as a plug-in module, augmenting exploration without degrading the base policy performance. Moreover, the structured latent space enables human-guided exploration, further improving efficiency and controllability. Extensive experiments in both simulation and real-world tasks demonstrate that SOE consistently outperforms prior methods, achieving higher task success rates, smoother and safer exploration, and superior sample efficiency. These results establish on-manifold exploration as a principled approach to sample-efficient policy self-improvement. 

\end{abstract}

\section{Introduction}

\emph{``We want AI agents that can discover like we can, not which contain what we have discovered."\\ \hspace*{35mm}--- Richard Sutton, The Bitter Lesson}\vspace{1mm}

In recent years, data-driven robot learning~\cite{chi2023diffusion, zhao2023learning, black2024pi_0, brohan2022rt, kim2024openvla} has attracted considerable attention, particularly for its potential to enhance robotic manipulation capabilities through large-scale data collection and training. By modeling visuomotor behaviors with neural networks, these approaches allow robot policies to learn from expert demonstrations and achieve near-human performance across a variety of tasks.

Despite these advances, most existing methods still rely heavily on human teleoperation for data acquisition~\cite{zhao2023learning, fang2025airexo} and policy refinement~\cite{liu2022robot, luo2025precise}, which presents several challenges. A primary concern is the high cost of teleoperation, as it typically requires skilled operators and specialized equipment, thereby limiting the scalability of data collection. More critically, teleoperated demonstrations often fail to cover the diverse scenarios a robot could encounter in the real world, resulting in distributional bias~\cite{zhang2025scizor} and compounding error~\cite{ross2011reduction}. The problem is further exacerbated by the fact that human operators may act based on contextual cues inaccessible to robot sensors. Robots, on the other hand, may internalize human habits rather than task-relevant behaviors. As a result, simply scaling up teleoperated data is not the optimal path toward improving policy performance.

Instead of passively imitating human-provided behavior, a line of research addresses this challenge by enabling robot policy self-improvement~\cite{bousmalis2023robocat, jin2025sime, mirchandani2024so, luo2024serl}---actively exploring the environment to collect diverse experience and leveraging that experience to refine policies. Under this paradigm, robots can autonomously discover novel behaviors that go beyond the coverage of human demonstrations. By iteratively practicing the learned behaviors, 
they also develop a deeper understanding of the natural variability in their actions, ultimately leading to a more robust and resilient policy.

The key to sample-efficient robot policy self-improvement lies in effective exploration. Prior work~\cite{ankile2024imitation, jin2025sime} has shown that imitation-learned policies often overfit demonstrations, collapse into single-modal motions, and fail to produce diverse behaviors. Without proper exploration, these policies tend to repeat failed behaviors, limiting their ability to discover improved solutions. While random exploration strategies can occasionally yield novel behaviors~\cite{lillicrap2015continuous}, they are generally ineffective in high-dimensional action spaces~\cite{li2025reinforcement} and can pose safety risks in real-world deployment~\cite{gu2024review}, causing potential hardware damage. This necessitates a more structured approach to exploration---one that ensures safety and effectiveness without sacrificing the diversity of experiences.

\begin{figure}
    \centering
    \includegraphics[width=\linewidth]{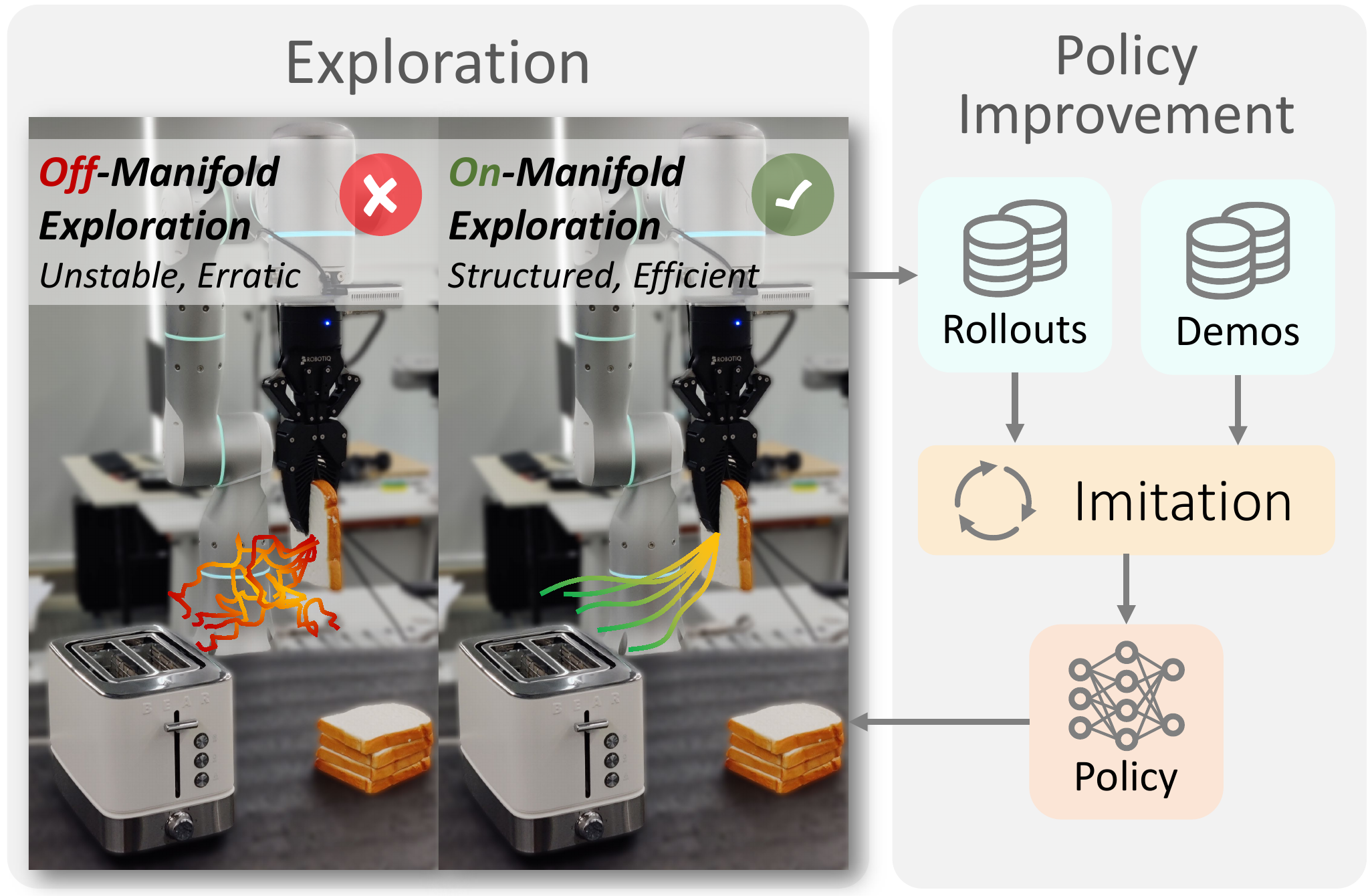}
    \caption{\textbf{Overview of SOE.} By constraining exploration to the manifold of valid actions, our approach generates diverse yet temporally coherent behaviors, enabling structured and efficient exploration. The collected rollout data is used to refine the policy, leading to efficient self-improvement.}
    \vspace{-6mm}
    \label{fig:teaser}
\end{figure}

To this end, we propose \textit{SOE}, a novel framework for \textit{Sample-Efficient Robot Policy \textbf{S}elf-improvement via \textbf{O}n-Manifold \textbf{E}xploration}. The core idea of our method is to ensure that exploration remains constrained to the manifold of valid actions---critical for both safety and effectiveness. Prior works often perturb the action space directly~\cite{lillicrap2015continuous} or inject random noise~\cite{jin2025sime}, leading to temporally inconsistent and unsafe behaviors, particularly under ``action chunking" representations~\cite{zhao2023learning}. In contrast, we perform exploration in a compact latent space learned through a variational information bottleneck (VIB). The latent representation in this space preserves only task-essential information in observation while discarding irrelevant details, ensuring exploration remains structured and efficient. As illustrated in Fig.~\ref{fig:teaser}, by operating on this latent representation, our framework enables effective on-manifold exploration and more robust policy improvement. Furthermore, we demonstrate that in the latent space, action chunks are naturally disentangled into distinct modes. Leveraging this property, we achieve controllable exploration, which allows users to guide exploration toward preferred directions, thereby enhancing interpretability and further boosting sample efficiency. Implemented as a plug-in module, our approach can be seamlessly integrated with existing imitation learning algorithms and jointly optimized, without any degradation in their performance.

To evaluate the effectiveness of our method, we conduct extensive experiments across a variety of robot manipulation tasks in both simulation and real world. The results show that \textit{SOE} consistently outperforms prior exploration methods in effectiveness, motion smoothness, and sample efficiency. With just one round of policy self-improvement, our method achieves substantial gains over the base policy, including an average relative improvement of 50.8\% on real-world tasks. Additional experiments in simulation and ablation studies further confirm multi-round performance improvements and the contribution of each component in our framework. Collectively, these findings demonstrate that on-manifold exploration provides a structured, safe, and effective approach to sample-efficient robot policy self-improvement.

\section{Related Work}


\subsection{Imitation Learning for Robot Manipulation}
Imitation learning is an extensively studied approach for training robot policies by mimicking expert demonstrations. Early methods, such as behavior cloning, directly learn a mapping from observations to actions using supervised learning. More recent approaches leverage advanced architectures, including transformers~\cite{zhao2023learning, shafiullah2022behavior, brohan2022rt, zitkovich2023rt, haldar2024baku} and flow-based generative models~\cite{janner2022planning, chi2023diffusion, ajay2022conditional, ze20243d, wang2024rise}, to capture multimodal visuomotor behaviors. These methods have demonstrated impressive performance on diverse manipulation tasks, promising a data-driven path for general-purpose robot learning~\cite{kim2024openvla, black2024pi_0, bjorck2025gr00t, team2024octo}. However, most of them still rely heavily on costly expert demonstrations, suffering from limited or biased data coverage, motivating the need for policy post-training.

\subsection{Real-World Post-training of Robot Policy}
Reinforcement learning (RL)~\cite{sutton1999policy, haarnoja2018soft, schulman2017proximal, lillicrap2015continuous} is a widely used paradigm for post-training robot policies, enabling robots to learn from trial and error. However, RL typically requires a prohibitive amount of interactions to achieve satisfactory performance, limiting their practicality for real-world applications. To address this, researchers have explored sim-to-real transfer~\cite{ankile2024imitation,lv2023sam}, offline RL~\cite{fujimoto2019off, kumar2020conservative, kostrikov2021offline}, residual policy learning~\cite{yuan2024policy, haldar2023teach}, and human-in-the-loop approaches~\cite{luo2025precise, chen2025conrft} to enhance sample efficiency. While effective in some cases, these methods face issues like extrapolation errors~\cite{kostrikov2021offline}, safety concerns~\cite{gu2024review}, and reliance on hand-crafted reward designs or labor-intensive human teleoperation. An alternative is to enable robot self-improvement through autonomous imitation learning~\cite{mirchandani2024so, bousmalis2023robocat, jin2025sime, liu2022robot}. However, imitation-learned policies often lack behavioral diversity, highlighting the need for effective exploration.

\subsection{Exploration in Robot Learning}
Exploration is crucial yet challenging in robot learning, particularly in high-dimensional spaces. Most existing exploration approaches can be categorized into two lines: reward-based methods and sampling-based methods. The first line of methods~\cite{parisi2021interesting, ecoffet2019go, burda2018large, pathak2017curiosity, kim2018emi, goyal2019infobot} achieves exploration by introducing some intrinsic rewards, based on novelty or curiosity, to incentivize the agent to visit unfamiliar or unpredictable states. The second line of methods diversifies action generation through strategies like epsilon-greedy~\cite{mnih2013playing, van2016deep}, Boltzmann exploration~\cite{sutton2005reinforcement, szepesvari2022algorithms}, and goal-directed exploration~\cite{hu2023planning}, which are typically achieved by perturbing the action space or injecting noise into the policy. Despite the success of both lines of methods in various domains, they often struggle in real-world manipulation with continuous, chunked actions. Perhaps the most relevant work to ours is \textit{SIME}~\cite{jin2025sime}, which perturbs diffusion policy conditions to induce modal-level exploration. Our method extends this idea by constraining exploration to the task manifold, enabling safer, more effective, and controllable exploration behaviors.

\section{Preliminaries and Problem Formulation}

In this section, we introduce the background of imitation learning and policy self-improvement. 

\textbf{Imitation Learning.}
Consider a robot manipulation task $\mathcal{T}$ along with a set of expert demonstrations $\mathcal{D}^e = \{\tau_i\}_{i=1}^N$, 
where each trajectory $\tau_i = (o_1, a_1, \ldots, o_T, a_T)$ consists of a sequence of observations $o_t \in \mathcal{O}$ and actions $a_t \in \mathcal{A}$. The goal of imitation learning is to learn a policy $\pi_\theta:\mathcal{O}\to\mathcal{A}$ that maps observations to actions by mimicking the expert behavior. This is typically achieved through 
optimizing the policy parameters $\theta$ by maximizing the likelihood of the expert actions given the corresponding observations:
\begin{equation}
    \label{eq:loss_imitate}
    \theta^* = \argmax_\theta \mathbb{E}_{(o_t,a_t)\sim\mathcal{D}^e}\left[\log \pi_\theta(a_t|o_t)\right].    
\end{equation}

\textbf{Policy Self-Improvement.}
Beyond imitation learning, policy self-improvement aims to enhance the learned policy by enabling the robot to actively explore its environment and collect additional experience. By learning from those experiences, the robot is expected to improve its performance beyond the limitations of the initial expert demonstrations. This procedure also aligns with the rejection sampling fine-tuning (RFT)~\cite{yuan2023scaling} paradigm in the large language model post-training literature, which has been proven effective in boosting model performance.

Formally, given the expert demonstration dataset $\mathcal{D}^e$, an imitation-learned policy $\pi_0$, potentially augmented with exploration mechanisms, interacts with the environment to collect a set of additional trajectories $\mathcal{D}^b$, 
which is merged with the original dataset to form an aggregated dataset $\mathcal{D} = \mathcal{D}^e \cup \mathcal{D}^b$. 
The policy is then refined on this aggregated dataset to obtain an improved policy $\pi_1$. This process can be iteratively repeated, yielding a sequence of progressively enhanced policies: $\pi_0, \pi_1, \pi_2, \ldots, \pi_m$.

The objective of policy self-improvement is to maximize the success rate of the policy $\pi_m$ while minimizing the number of interactions $|\mathcal{D}^b|$ required to achieve this improvement and ensuring the safety of the exploration process.

\section{Method}

In this paper, we present \textit{SOE}, a novel framework for sample-efficient robot policy self-improvement via on-manifold exploration. We begin by outlining the challenges of exploration in high-dimensional action spaces and introduce on-manifold exploration, which involves learning a compact latent representation that captures task-essential factors. Next, we describe how our exploration mechanism can be seamlessly integrated into existing imitation learning pipelines as a plug-in module, without degrading base policy performance. Finally, we introduce user-guided steering, which leverages the disentangled nature of the learned representation for controllable exploration.






\subsection{On-Manifold Exploration}

Exploration is critical for policy self-improvement, as it enables the robot to discover novel behaviors beyond the coverage of expert demonstrations. However, achieving effective exploration in action spaces is notoriously challenging, particularly when actions are continuous and temporally grouped in chunks. This is because the action space is typically vast, and the task-relevant manifold only occupies a small fraction of it. Directly injecting random noise into actions often drives the policy off the task manifold, leading to unsafe, jerky motions and ineffective exploration, as illustrated in Fig.~\ref{fig:teaser}. Perturbing latent observation embeddings may appear to be an alternative, but since these embeddings are usually entangled with redundant or correlated features, such perturbations can still yield unrealistic or out-of-distribution actions, as demonstrated in our experiments.

This motivates the need for learning a more structured, well-shaped feature space, which can effectively capture the underlying task manifold and facilitate sample-efficient on-manifold exploration. To address this, we propose learning a compact latent representation of observations that retains only the information essential for policy execution while discarding irrelevant, high-frequency details. This yields a more structured and robust target for perturbations, enabling effective on-manifold exploration. Formally, given observation $O$ and action $A$, we define the latent variable $Z$ via an encoder $p_\theta(z|o)$, optimized with the following objective:
\begin{equation}
    \max_\theta I(Z;A)-\beta I(Z;O),\label{eq:info}
\end{equation}
where $I(\cdot;\cdot)$ denotes mutual information and $\beta$ controls the trade-off between informativeness and compactness. This encourages $Z$ to preserve action-relevant features while minimizing dependence on extraneous observation details. Following the variational information bottleneck (VIB)~\cite{alemi2016deep}, we derive a tractable variational upper bound from Eqn.~\ref{eq:info}:
\begin{align}
\label{eq:vib_loss_org}
&\mathcal{L}_\text{IB}(\theta,\phi) = \nonumber\\ &\ \ \mathbb{E}_{(o,a)\sim\mathcal{D}}[-\mathbb{E}_{z\sim p_\theta(z|o)}\log q_\phi(a|z)+\beta\text{KL}[p_\theta(Z|o)||r(Z)]],
\end{align}
where $q_\phi(a|z)$ is a decoder that reconstructs actions from $Z$, and $r(Z)$ is a predefined prior. The first term, denoted as $\mathcal{L}_{\text{IB(IL)}}$, corresponds to the imitation loss in Eqn.~\ref{eq:loss_imitate}, encouraging $Z$ to retain action-relevant information. The second term, denoted as $\mathcal{L}_{\text{IB(KL)}}$, acts as a regularizer that penalizes deviations from the prior. In our implementation, the encoder $p_\theta$ is parameterized as a diagonal Gaussian model, and the prior $r(Z)$ is chosen as a $d$-dimensional isotropic Gaussian $\mathcal{N}(Z;0,I)$.

\begin{figure}[t]
    \centering
    \includegraphics[width=\linewidth]{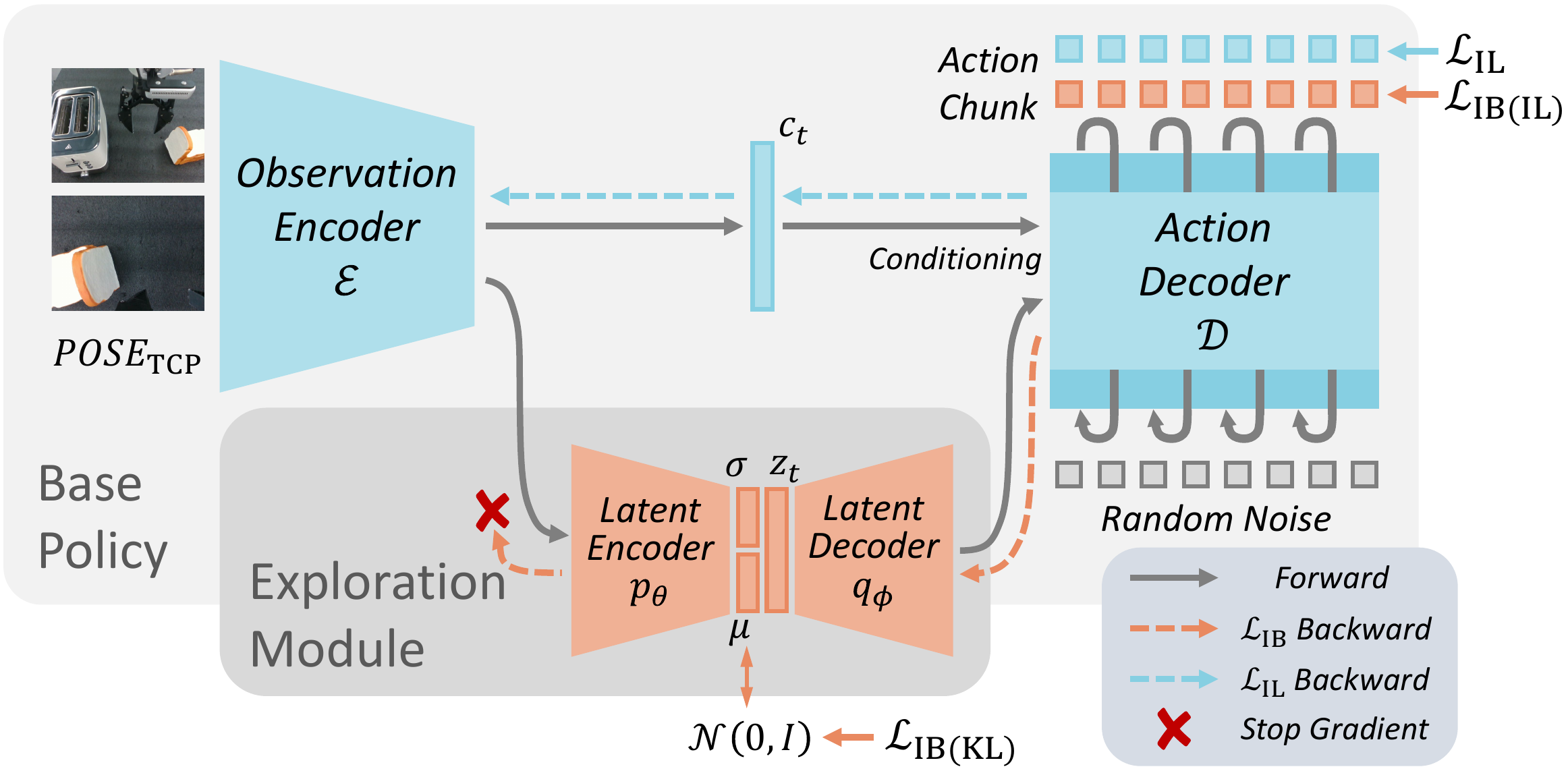}
    \caption{\textbf{Dual-Path Architecture of \textit{SOE}.} The observation embedding is processed through two parallel paths: the base path (top), responsible for stable policy execution, and the exploration path (bottom), responsible for generating diverse actions. Each path outputs distinct noise predictions and is optimized with a separate loss function. }
    \vspace{-5mm}
    \label{fig:pipeline}
\end{figure}

With the latent representation $Z$, the underlying low-dimensional task manifold is explicitly modeled. By sampling from $p_\theta$ and decoding through $q_\phi$, we can generate diverse, on-manifold action proposals:
\begin{gather*}
\mu_t,\sigma_t \sim p_\theta(Z|O=o_t),\quad
z_t \sim \mathcal{N}(\mu_t,(\alpha\sigma_t)^2),\nonumber\\
a_t \sim q_\phi(A|Z=z_t),
\end{gather*}
where $\mu_t$ and $\sigma_t$ establish the local geometry of the task manifold around the current observation $o_t$, and $\alpha$ is a hyperparameter that controls the exploration scale. A larger $\alpha$ expands exploration to a broader neighborhood on the manifold, while a smaller $\alpha$ restricts exploration to local variations. By adjusting $\alpha$, the exploration strategy can smoothly transition between conservative and aggressive, providing a flexible mechanism to balance safety and diversity.

Compared to na\"ive random noise injection, our approach leverages the structured distribution of the learned latent space to perform exploration via sampling. This ensures that the generated action proposals remain on the task manifold, avoiding unrealistic or unsafe behaviors, while still supporting diverse and informative exploration, as illustrated in Fig.~\ref{fig:teaser}. With this on-manifold exploration mechanism, the robot can efficiently discover successful behaviors, enabling sample-efficient robot policy self-improvement.

\subsection{Exploration as a Plug-in}

To enhance the general applicability of the proposed approach, a central question is how to incorporate the exploration mechanism into standard imitation learning pipelines. Two challenges are particularly salient. First, the integration must preserve the expressive capacity of base policy, ensuring exploration does not compromise its fundamental performance. Second, the introduced component should impose minimal computational and optimization overhead, ideally allowing joint, end-to-end training with base policy.

We address these challenges by designing our exploration mechanism as a plug-in module that can be seamlessly integrated and jointly optimized with existing policy networks. As illustrated in Fig.~\ref{fig:pipeline}, this is achieved through a dual-path architecture that supports both stable policy execution and diverse exploration. In this setup, the latent encoder $p_\theta$ and decoder $q_\phi$ form an auxiliary bypass alongside the original observation encoder $\mathcal{E}$ and action decoder $\mathcal{D}$ of the base policy, resulting in two parallel action-generation paths:
\begin{itemize}
    \item \textbf{Base path}, where the embedding from the observation encoder $\mathcal{E}$ directly conditions the action decoder $\mathcal{D}$:
\begin{equation*}
    \label{eq:base}
    c_t = \mathcal{E}(o_t),\quad a_{t:t+H} = \mathcal{D}(c_t),
\end{equation*}
where $c_t$ is the observation embedding at time $t$ and $a_{t:t+H}$ is the generated action chunk spanning $H$ steps.

    \item \textbf{Exploration path}, where the observation embedding $c_t$ is encoded into the latent space, perturbed stochastically, and then decoded back into a modified embedding $\tilde{c}_t$ for diverse action proposal generation:
\begin{gather*}
    c_t = \mathcal{E}(o_t),\quad\mu_t,\sigma_t = p_\theta(c_t),\quad z_t \sim \mathcal{N}(\mu_t,(\alpha\sigma_t)^2),\nonumber\\
    \tilde{c}_t=q_\phi(z_t),\quad \tilde{a}_{t:t+H} = \mathcal{D}(\tilde{c}_t).
\end{gather*}
\end{itemize}

The base path ensures that the core policy remains intact, while the exploration path facilitates the generation of varied action proposals through structured perturbations in the latent space. Together, they form a switchable exploration mechanism, allowing users to alternate between standard execution and active exploration as needed.

During training, both paths are optimized jointly. The base path is trained with a standard imitation loss $\mathcal{L}_\text{IL}$ (Eqn.~\ref{eq:loss_imitate}), while the exploration path is trained with a variational information bottleneck loss $\mathcal{L}_\text{IB}$ (Eqn.~\ref{eq:vib_loss_org}), which combines action reconstruction with KL regularization. We instantiate the base policy as a diffusion policy~\cite{chi2023diffusion}, which leads to the following losses:
\begin{gather}
    \mathcal{L}_\text{IL}(\psi) = \mathbb{E}_{\substack{(o,a)\sim\mathcal{D},\epsilon\sim\mathcal{N}(0,I)\\k\sim\text{Uniform}\{1,\ldots,K\}}}
    \left[||\epsilon - \epsilon_\psi(a^k_{t:t+H},c_t,k)||^2\right],
\end{gather}
\begin{align}
    \mathcal{L}_\text{IB}(\theta,\phi) = \mathbb{E}_{(o,a)\sim\mathcal{D}}&\left[\mathbb{E}_{\epsilon, k}\left[||\epsilon - \epsilon_\psi(a^k_{t:t+H},\tilde{c}_t,k)||^2\right]\right.\nonumber\\
    \label{eq:vib_loss}
    &\ \left.+\beta\text{KL}[p_\theta(Z|o)||r(Z)]\right],
\end{align}
where $\epsilon_\psi$ is the diffusion policy parameterized by $\psi$, and $a^k_{t:t+H}$ is the noisy action at diffusion step $k$. The overall training objective is a combination of both losses:
\begin{equation}
    \label{eq:loss}
    \mathcal{L}(\theta,\phi,\psi) = \mathcal{L}_\text{IL}(\psi) + \mathcal{L}_\text{IB}(\theta,\phi).
\end{equation}

Notably, $\mathcal{L}_\text{IB}(\theta,\phi)$ updates only $p_\theta$ and $q_\phi$, leaving the base policy unaffected, while $\mathcal{L}_\text{IL}(\psi)$ updates only $\mathcal{E}$ and $\mathcal{D}$. This design ensures that the exploration mechanism does not come at the cost of degrading base policy performance. Furthermore, by offloading action-level multimodal modeling to the diffusion process, the latent encoder can focus on capturing cognitive-level uncertainty, leading to a more robust characterization of the underlying task manifold.

\subsection{User-Guided Steering}

In addition to facilitating on-manifold exploration, the learned latent representation $Z$ also provides a natural mechanism for controllable exploration. Owing to the disentanglement encouraged by $\mathcal{L}_{\text{IB}}$ (Eqn.~\ref{eq:vib_loss}), different dimensions of $Z$ tend to correspond to distinct task-relevant factors. For example, in a cup-grasping task, one dimension may encode the cup’s horizontal position, while another captures its vertical position. Such a disentangled structure allows users to intuitively guide exploration by perturbing specific latent dimensions, steering the policy toward desired behaviors.

We achieve this by first identifying the most informative dimensions of $Z$ using a signal-to-noise ratio (SNR) criterion, and then restricting steering to these dimensions. The SNR is defined as:
\begin{equation*}
    \text{SNR}_i = \frac{\text{Var}(\mu_i)}{\mathbb{E}[\sigma_i^2]},\quad i=1,2,\dots,d,
\end{equation*}
where $\mu_i$ and $\sigma_i$ are the mean and standard deviation of the $i$-th dimension of $Z$. A high SNR indicates the dimension encodes reliable, task-relevant information, while a low SNR suggests it is uninformative or even collapsed, as validated in our experiments Sec.~\ref{sec:realworld}. There is also a clear separation between effective and ineffective dimensions, allowing us to define a universal, task-agnostic threshold for categorization.

Identifying the effective dimensions substantially reduces the search space for steering, allowing users to concentrate on the most relevant factors. Furthermore, to ensure broad coverage of the task manifold, we generate a large batch of action proposals for each effective dimension and apply farthest point sampling (FPS) to select a diverse subset for user selection. By starting with a large batch size and retaining only a small number of representative actions, the selected proposals effectively capture the full range of variations along each dimension while avoiding redundancy.

These proposals are then presented to users through an interactive interface, enabling them to explore the activated latent dimensions and inspect the corresponding behaviors. By choosing preferred actions, users can directly steer the robot toward desired behaviors. This design introduces human-in-the-loop guidance during exploration, achieving even higher sample efficiency than autonomous self-improvement, while eliminating the need for specialized teleoperation devices or the exhausting effort of manual teleoperation.

\section{Experiments}

In this section, we evaluate our method on a range of robot manipulation tasks in both simulation and real-world settings. We start by introducing the setups for our experiments.

\subsection{Experiment Setups}

\subsubsection{Real World}
We deploy our approach on a Flexiv Rizon4 robotic arm with a Robotiq 2F-85 gripper. We attach soft fingers to the grippers following~\cite{chi2024universal}. The visual observations are provided by two Intel RealSense D435i cameras: one fixed on the side and one mounted on the wrist.

The robot is controlled in Cartesian space at 10 Hz. Each action chunk spans $H=20$ steps and is represented in relative poses. 
The action space is 10-dimensional, comprising 3D translation, 6D rotation, and 1D gripper control, while the observation space includes RGB images ($216\times288\times3$) from both cameras, 
along with proprioception states. 

\begin{figure}[htbp]
    \centering
    \includegraphics[width=\linewidth]{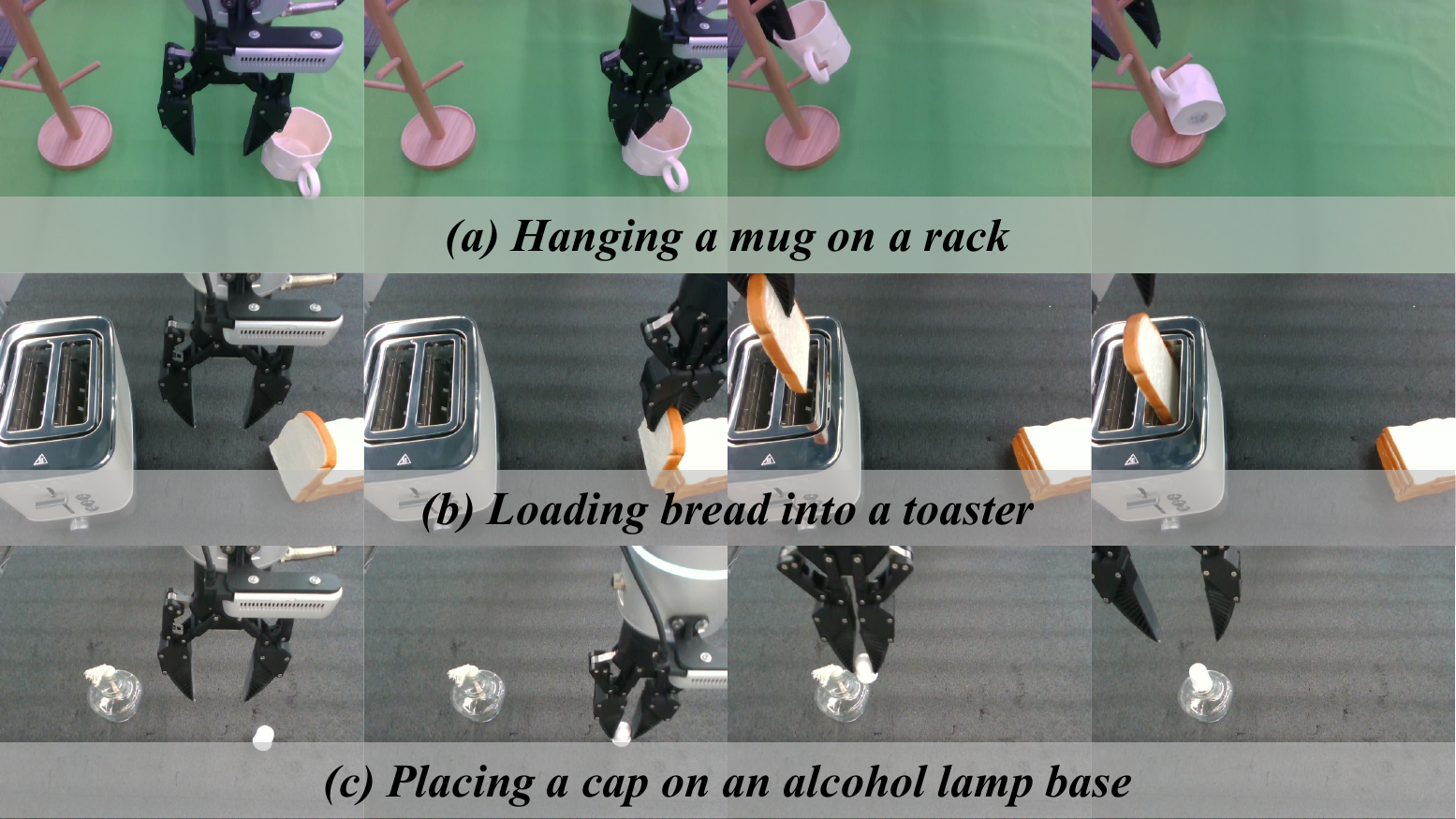}
    \caption{\textbf{Overview of real-world manipulation tasks}, which include (a) \textit{Mug Hang}, (b) \textit{Toaster Load}, and (c) \textit{Lamp Cap}.}
    \label{fig:tasks}
\end{figure}

We evaluate our method on three real-world manipulation tasks: \textit{Mug Hang} (grasp a mug and hang it on a rack), \textit{Toaster Load} (pick up a slice of bread and insert it into a toaster), and \textit{Lamp Cap} (assemble an alcohol lamp by placing its cap onto its base). Fig.~\ref{fig:tasks} provides an illustration of all three tasks.

For each task, we collect a limited set of human teleoperated demonstrations using a sigma.7 haptic device: 30 demonstrations each for \textit{Toaster Load} and \textit{Lamp Cap}, and 50 for \textit{Mug Hang}, as it is relatively more complex. These demonstrations are used to train the initial base policy and serve as a foundation for subsequent self-improvement.

\subsubsection{Simulation}

We also evaluate our method on a set of simulation tasks from RoboMimic~\cite{mandlekar2021matters}. 
The tasks include \textit{Lift}, \textit{Can}, \textit{Square}, and \textit{Transport}. 
The observations are restricted to RGB images from a fixed agent-view camera and an in-hand camera, along with proprioceptive states.

We use the officially provided proficient-human (ph) demonstrations for policy initialization. Since Diffusion Policy already achieves near-perfect performance with the full set of 200 demonstrations~\cite{chi2023diffusion}, we consider a more challenging few-shot regime, following prior work~\cite{jin2025sime}. Specifically, we randomly sample 20 demonstrations for \textit{Can}, \textit{Square}, and \textit{Transport}, and 10 demonstrations for \textit{Lift}, given its relative simplicity. Starting from these limited demonstrations and the resulting imperfect base policies, the robot is expected to improve itself through active exploration.

\subsubsection{Metrics}
Our primary metric is \textit{task success rate}. 
To measure exploration effectiveness, we report \textit{Pass@5}, the probability of achieving at least one success within five attempts from the same starting condition. To reflect sample efficiency, we record the \textit{number of rollouts} required to collect these successful experiences. To assess safety and motion quality, we also compute the \textit{average jerk} of the end-effector trajectory, defined as
$J_\tau = \frac{1}{T}\sum_{t=1}^{T} ||\frac{d^3 p_t}{dt^3}||$,
where $p_t$ denotes the end-effector position at time $t$. A lower jerk value indicates smoother, safer motions, which is crucial for real-world deployment.

\subsubsection{Policy}
Our method is implemented on top of Diffusion Policy~\cite{chi2023diffusion} with ResNet-18~\cite{he2016deep} as the visual encoder and DDIM~\cite{song2020denoising} as the scheduler. For our plug-in exploration module, both the latent encoder and decoder are 3-layer MLPs with ReLU activations, and the latent dimension is set to $d=16$. 
Policies are optimized using AdamW with a learning rate of 3e-4, a batch size of 256, and trained for 25k iterations on two NVIDIA A800 GPUs. 

\subsubsection{Baselines}
We compare our method against two baselines: the basic \textit{Diffusion Policy (DP)}~\cite{chi2023diffusion} without explicit exploration mechanisms, and \textit{SIME}~\cite{jin2025sime}, a recently proposed approach that enables modal-level exploration by injecting random noise into diffusion conditioning. For a fair comparison, all methods are implemented to share the same base policy architecture and training recipe, differing only in their exploration mechanisms.

\subsection{Evaluation on Real-World Tasks}
\label{sec:realworld}

\begin{table*}[htbp]
    \centering
    \caption{Experiment Results on Real-World Tasks}
    \label{tab:realworld}
    \begin{tabular}{c|c|ccccc|c}
    \toprule
    \multicolumn{2}{c|}{} 
    & \multirow{2}{*}{\textbf{Pass@5} $\uparrow$}
    & \multirow{2}{*}{\textbf{Average Jerk} $\downarrow$}
    & \multirow{2}{*}{\textbf{Rollout Num} $\downarrow$}
    & \multicolumn{2}{c|}{\textbf{Success Rate $\uparrow$}}
    & \multirow{2}{*}{\textbf{Relative Improvement}}\\
    \multicolumn{2}{c|}{} 
    &&&
    &Before Imp. 
    &After Imp.
    &\\

    \midrule
    \multirow{4}{*}{\textit{Mug Hang}}
    & DP   
    & 0.56 & 3.34 & 77 & 0.47 & 0.38 & $-19.1\%$ \\
    & SIME 
    & 0.69 & 5.14 (+1.80) & 65 & 0.47 & 0.50 & $+6.4\%$  \\
    & SOE (Ours)
    & 0.75 & 3.57 (+0.23) & 60 & 0.47 & 0.56 & $+19.1\%$ \\
    & SOE + steering (Ours)
    & \textbf{0.81} & 3.68 (+0.34) & \textbf{53} & 0.47 & \textbf{0.66} & $+40.4\%$ \\

    \midrule
    \multirow{4}{*}{\textit{Toaster Load}}
    & DP   
    & 0.66 & 2.64 & 59 & 0.56 & 0.62 & $+10.7\%$ \\
    & SIME 
    & 0.84 & 3.64 (+1.00) & 51 & 0.56 & 0.62 & $+10.7\%$ \\
    & SOE (Ours)
    & 0.94 & 2.88 (+0.24) & 36 & 0.56 & 0.75 & $+33.9\%$ \\
    & SOE + steering (Ours)
    & \textbf{1.00} & 2.89 (+0.25) & \textbf{32} & 0.56 & \textbf{0.84} & $+50.0\%$ \\

    \midrule
    \multirow{4}{*}{\textit{Lamp Cap}}
    & DP   
    & 0.62 & 2.88 & 34 & 0.50 & 0.56 & $+12.0\%$ \\
    & SIME 
    & 0.69 & 3.53 (+0.65) & 36 & 0.50 & 0.50 & $0.0\%$ \\
    & SOE (Ours)
    & 0.88 & 2.97 (+0.09) & 30 & 0.50 & 0.69 & $+38.0\%$ \\
    & SOE + steering (Ours)
    & \textbf{0.94} & 2.90 (+0.02) & \textbf{25} & 0.50 & \textbf{0.81} & $+62.0\%$ \\
    
    \bottomrule
    \end{tabular}
\end{table*}

In this section, we aim to answer the following two questions: (1) Does \textit{SOE} lead to more effective, safer, and more efficient exploration compared to prior methods? (2) Can the learned latent representation capture task-relevant information and facilitate structured exploration as well as user-guided steering? 

\begin{figure}[t]
    \centering
    \begin{subfigure}{0.32\linewidth}
    \centering
    \includegraphics[width=\linewidth]{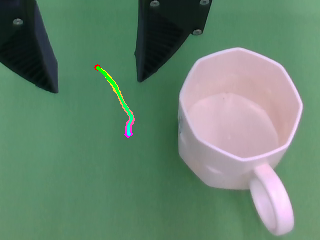}
    \vspace{-3mm}
    
    \includegraphics[width=\linewidth]{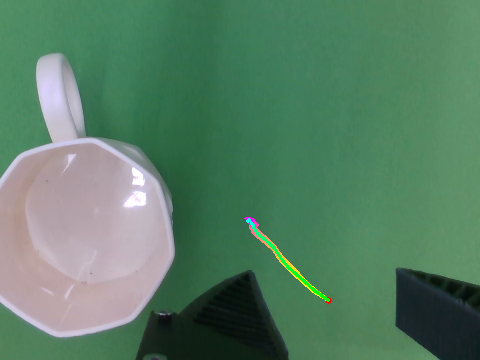}
    \caption{DP}
    \end{subfigure}
    \begin{subfigure}{0.32\linewidth}
    \centering
    \includegraphics[width=\linewidth]{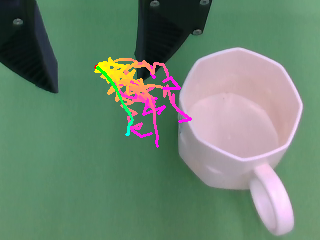}
    \vspace{-3mm}
    
    \includegraphics[width=\linewidth]{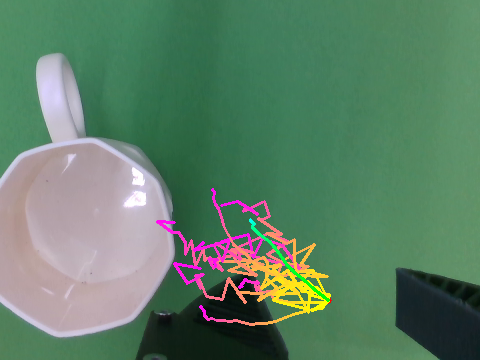}
    \caption{SIME}
    \end{subfigure}
    \begin{subfigure}{0.32\linewidth}
    \centering
    \includegraphics[width=\linewidth]{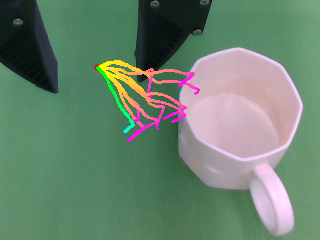}
    \vspace{-3mm}
    
    \includegraphics[width=\linewidth]{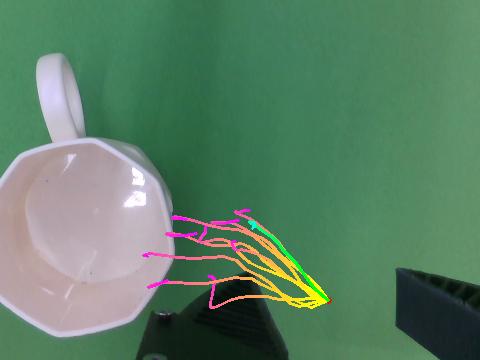}
    \caption{SOE (Ours)}
    \end{subfigure}
    
    \caption{\textbf{Comparison of action proposals from different methods.} \textit{DP} tends to generate repetitive failures, \textit{SIME} explores more broadly but with erratic, temporally inconsistent motions, whereas our method \textit{SOE} produces purposeful and structured action proposals.}
    \vspace{-4mm}
    \label{fig:comparison}
\end{figure}

As shown in Table~\ref{tab:realworld}, \textit{SOE} consistently outperforms all baselines across three tasks in terms of effectiveness, safety, and efficiency. The basic \textit{Diffusion Policy (DP)} exhibits low Pass@5 rates, often repeating similar unsuccessful behaviors, and can even suffer performance degradation after training on self-collected data. \textit{SIME} achieves higher Pass@5 but produces jerky and unsafe motions, limiting the quality of the collected data and resulting in weak improvement on those precision-demanding tasks. In contrast, our approach achieves significant improvements in success rate even without human-in-the-loop steering, while maintaining smooth, safe trajectories and requiring fewer rollouts. A qualitative comparison of action proposals from different methods is shown in Fig.~\ref{fig:comparison}. Together, these results underscore the benefits of on-manifold exploration in enhancing the effectiveness, safety, and efficiency of robot policy self-improvement. With user-guided steering, exploration can be further directed toward promising directions, resulting in even greater improvements with fewer rollouts.

\begin{figure}[htbp]
    \centering
    \begin{subfigure}{\linewidth}
        \centering
        \includegraphics[width=0.32\linewidth]{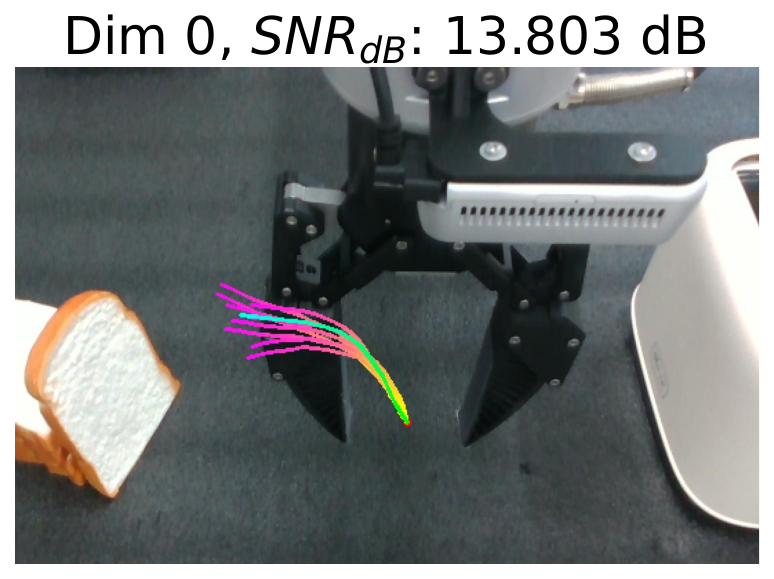}
        \includegraphics[width=0.32\linewidth]{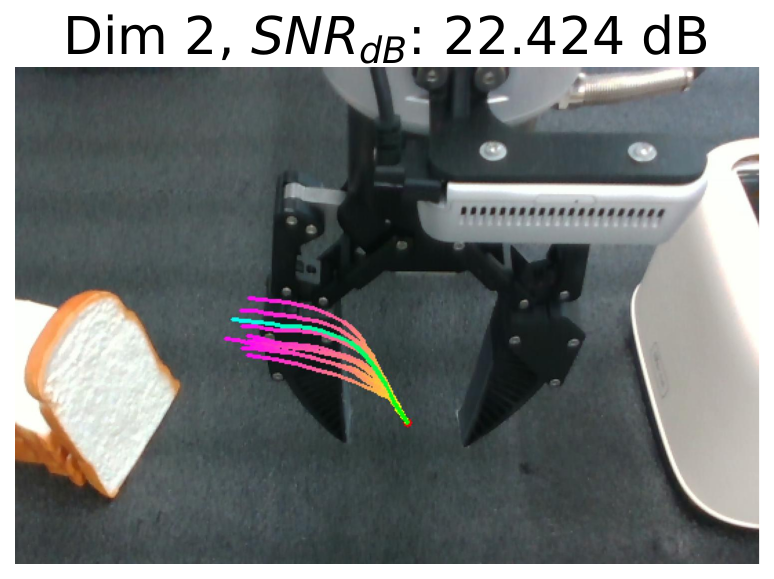}
        \includegraphics[width=0.32\linewidth]{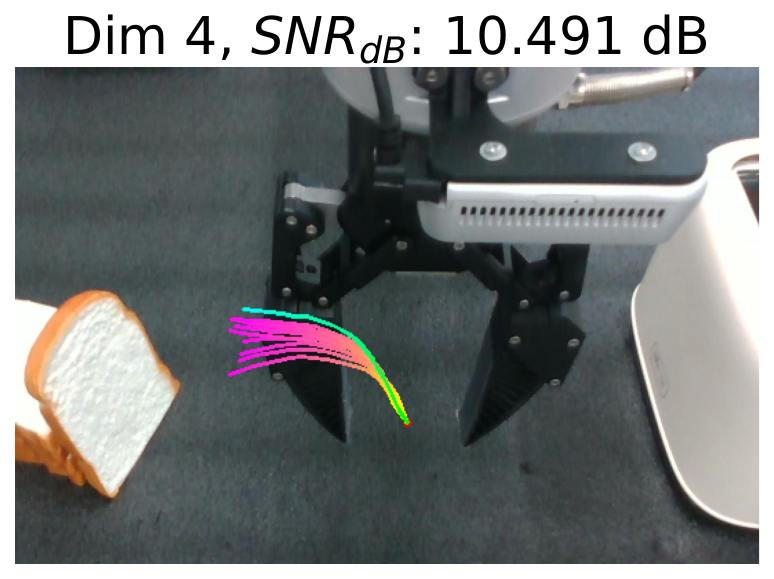}
    \end{subfigure}
    \newline
    \begin{subfigure}{\linewidth}
        \centering    
        \includegraphics[width=0.32\linewidth]{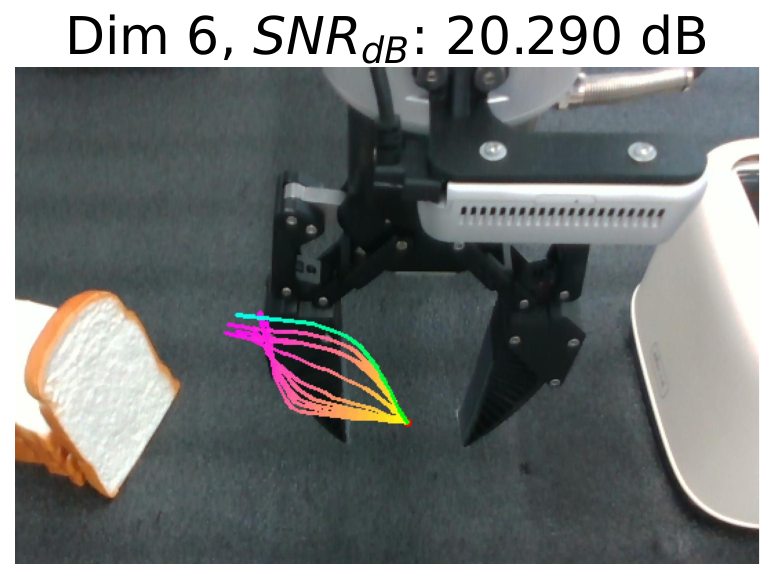}
        \includegraphics[width=0.32\linewidth]{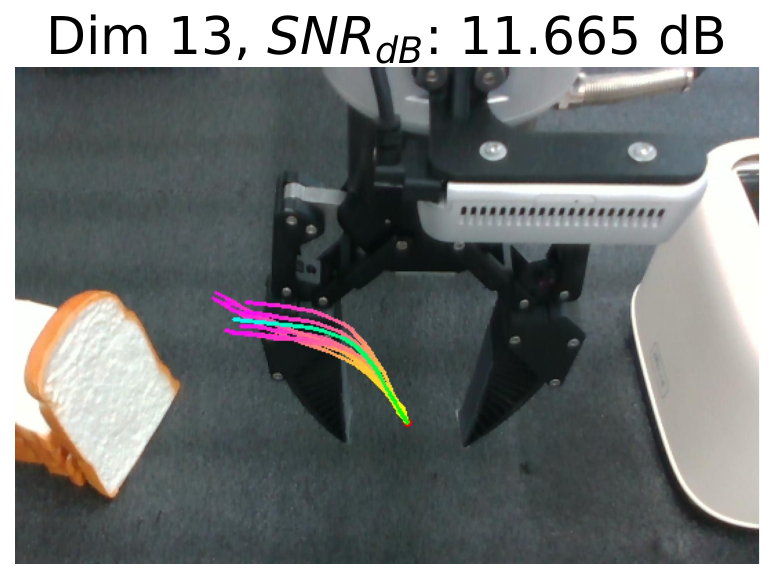}
        \includegraphics[width=0.32\linewidth]{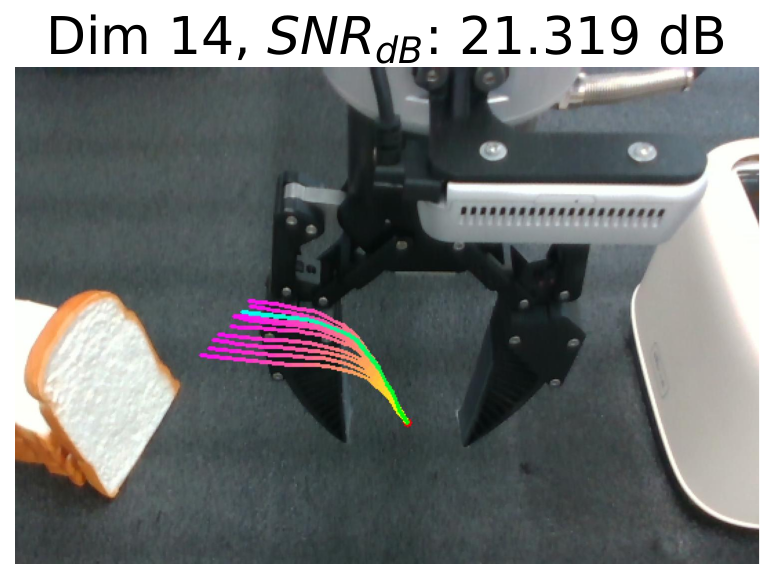}
    \end{subfigure}
    \newline
    \begin{subfigure}{\linewidth}
        \centering
        \includegraphics[width=0.32\linewidth]{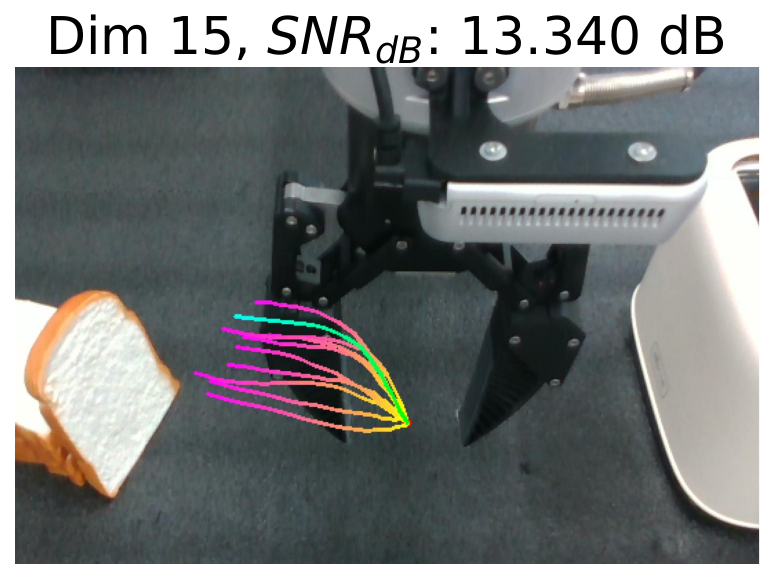}    
        \includegraphics[width=0.32\linewidth]{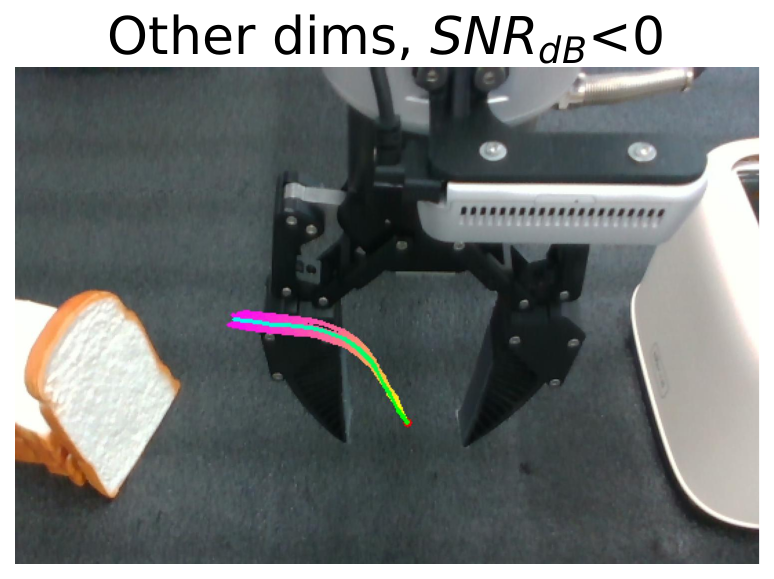}
    \end{subfigure}
    \caption{\textbf{Visualization of action proposals from different latent dimensions.} Each subfigure corresponds to a specific latent dimension, with the title indicating the dimension index and its SNR value (measured in dB). 
    Activating dimensions with positive SNR\textsubscript{dB} leads to meaningful and diverse patterns of action variation, while activating a low-SNR dimension (bottom-right) results in negligible diversity.}
    \vspace{-5mm}
    \label{fig:steering}
\end{figure}

To gain a deeper insight into the learned latent space, we visualize the action proposals used for user-guided steering in Fig.~\ref{fig:steering}. By activating one dimension at a time and plotting the resulting action proposals as trajectory points, we can observe how each dimension individually influences the robot's motion. As shown in the figure, different dimensions induce distinct patterns of action variation, such as moving left/right, up/down, or straight/curved, suggesting a certain degree of disentanglement in the latent space. Moreover, dimensions with high SNR---referred to as effective dimensions---produce meaningful trajectory variations when activated, while low-SNR dimensions yield negligible changes, indicating they contribute little to action generation. By focusing exploration and steering on the effective dimensions, our method can substantially reduce the search space, thereby resulting in more structured and efficient exploration.

\subsection{Evaluation on Simulation Benchmark}
\label{sec:sim}

\begin{figure*}[htbp]
    \centering
    \includegraphics[width=0.23\linewidth]{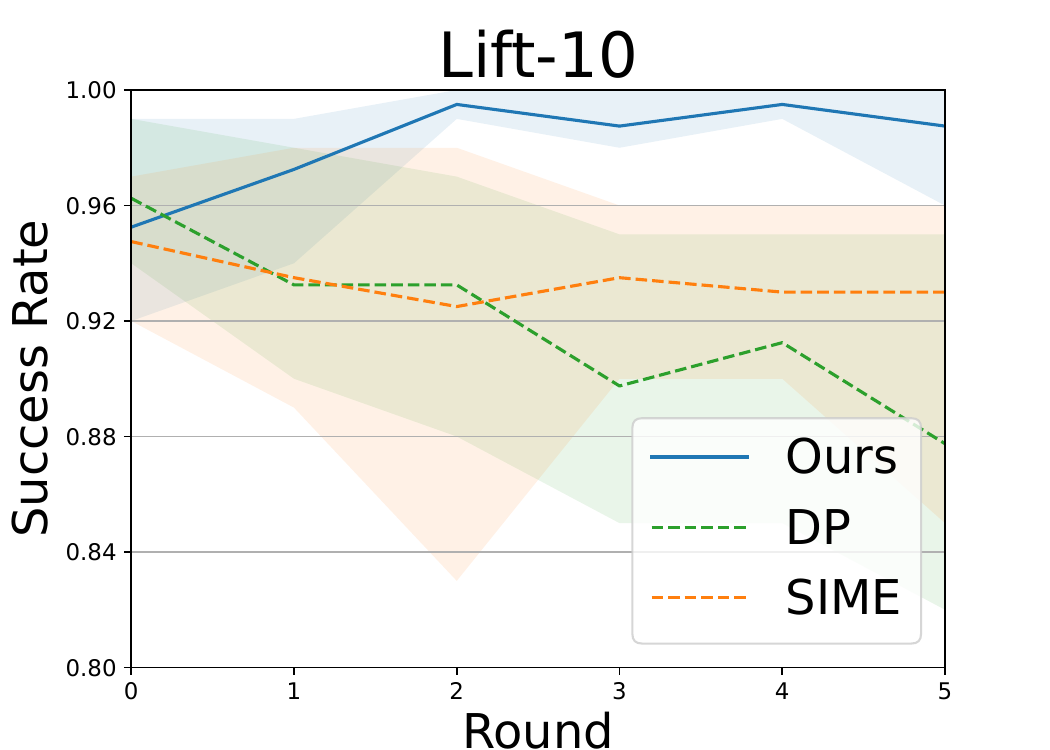}
    \includegraphics[width=0.23\linewidth]{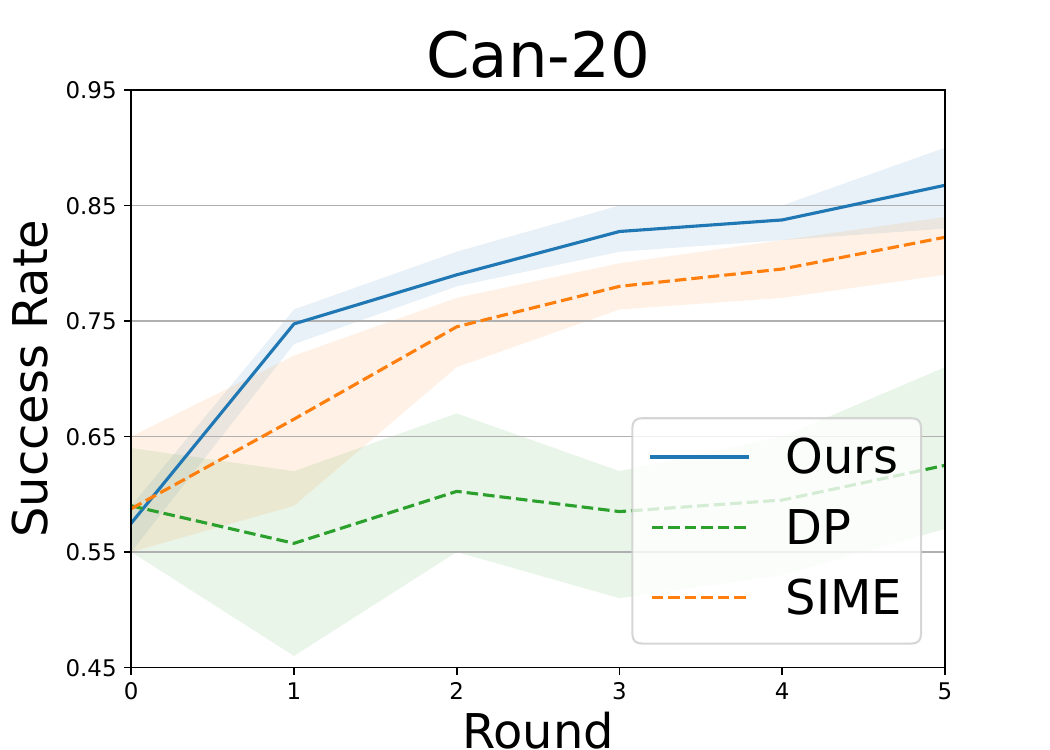}
    \includegraphics[width=0.23\linewidth]{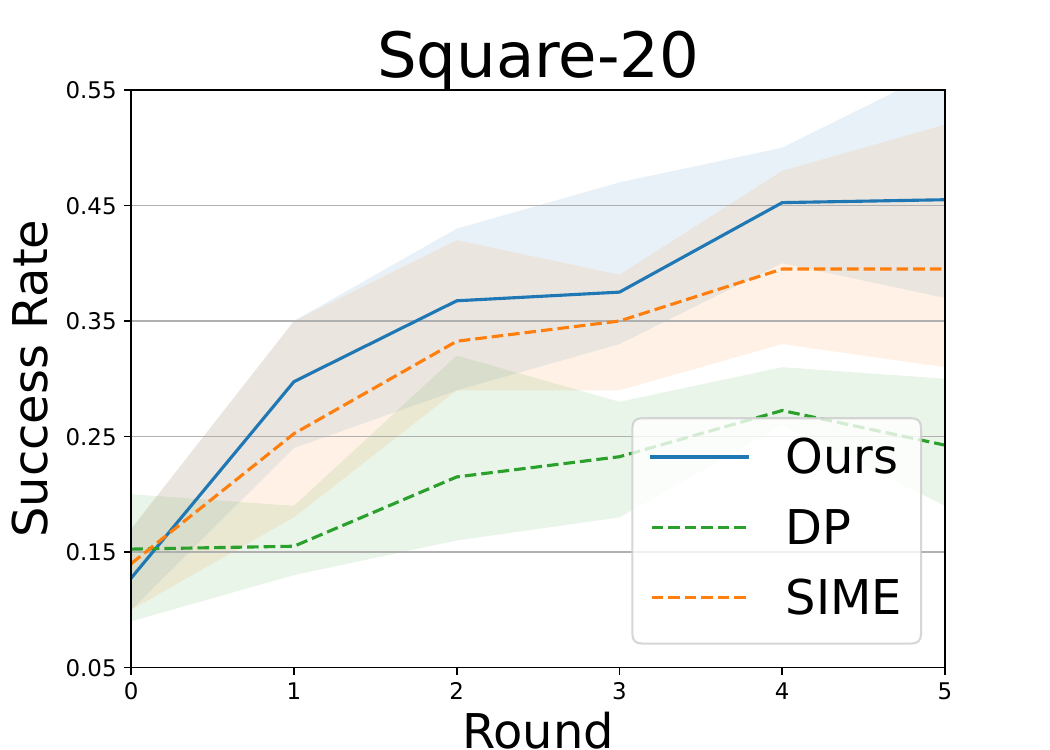}
    \includegraphics[width=0.23\linewidth]{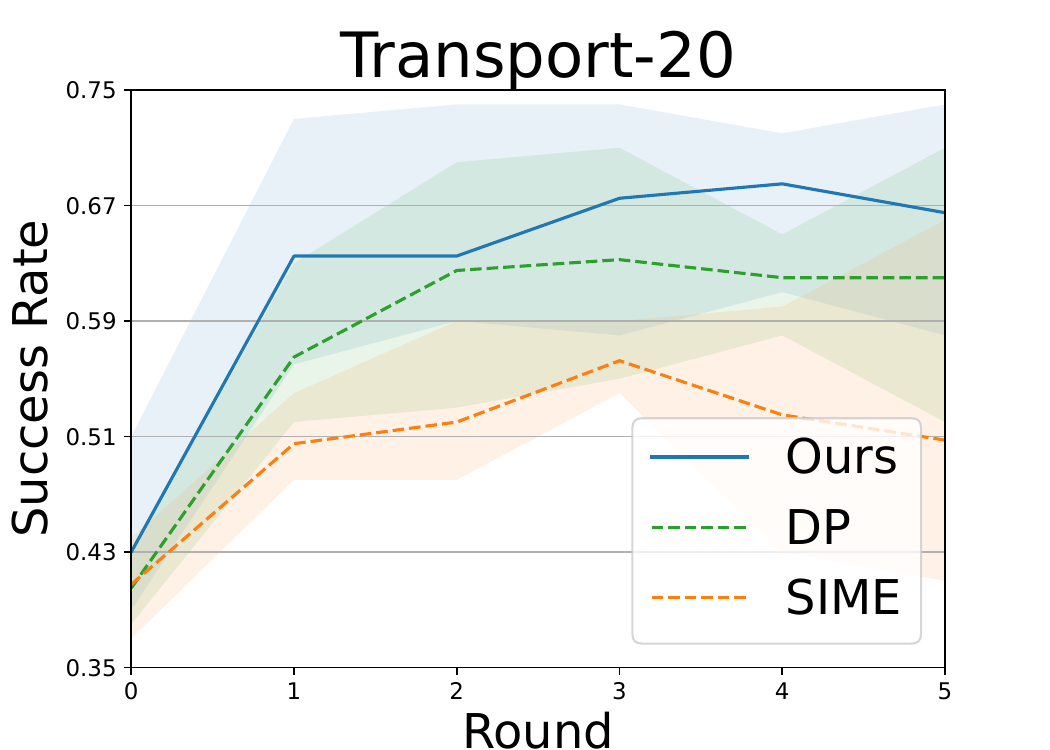}
    \caption{\textbf{Task success rate over rounds.} Our method consistently improves policies, whereas baselines perform unstably.}
    \vspace{-5mm}
    \label{fig:sim_sr}
\end{figure*}

\begin{figure}[htbp]
    \centering
    \includegraphics[width=0.48\linewidth]{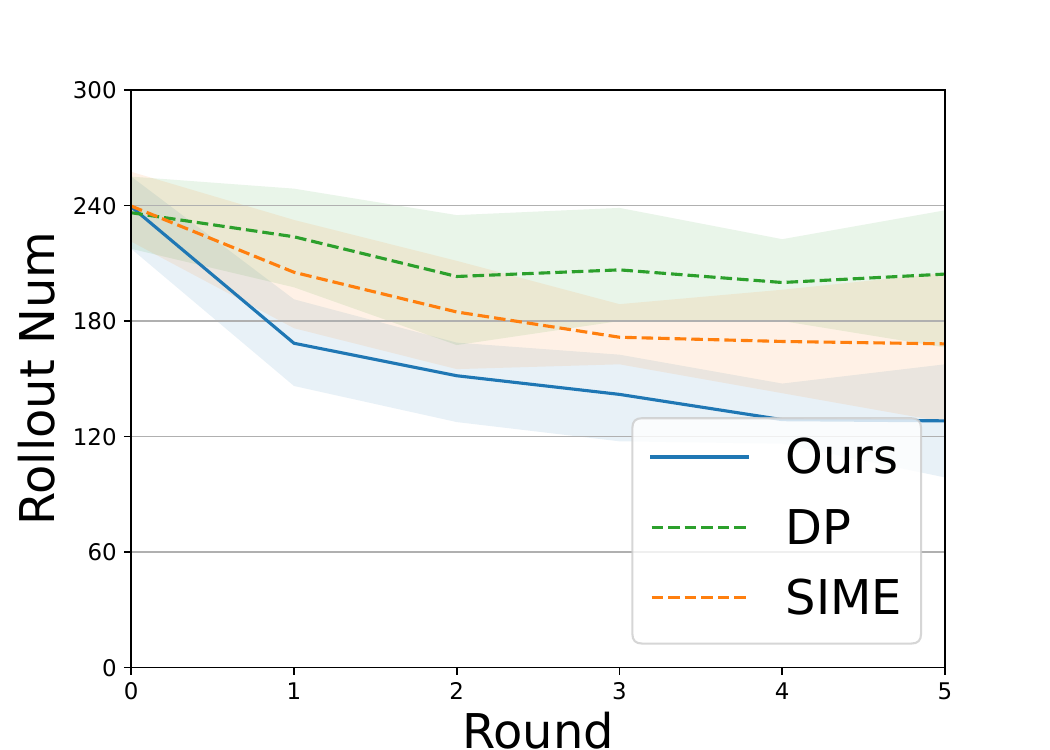}
    \includegraphics[width=0.48\linewidth]{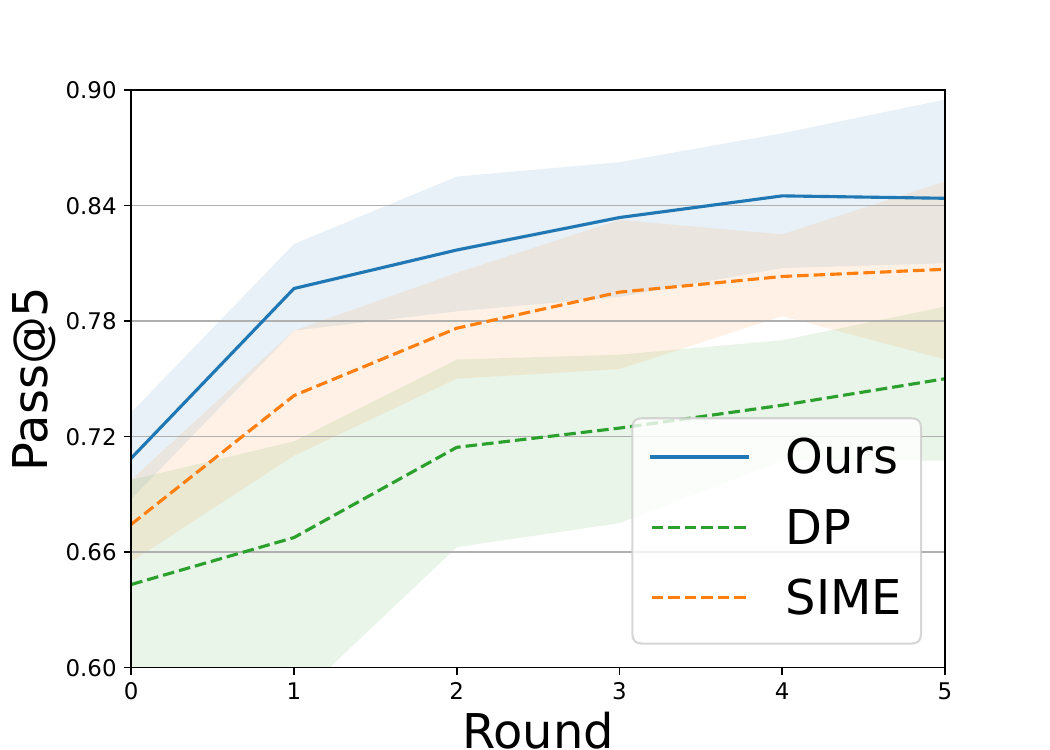}
    \caption{\textbf{Number of rollouts and exploration success rate over rounds, averaged across tasks.} Our method generally achieves higher Pass@5 rates while requiring fewer environment interactions compared to baselines, indicating more efficient and more effective exploration.}
    \label{fig:sim_exp}
    \vspace{-4mm}
\end{figure}

To further investigate the robustness and scalability of \textit{SOE}, we conduct a multi-round evaluation experiment on four vision-based manipulation tasks from RoboMimic. For each task, we train the policy under 4 different random seeds and evaluate it on 100 distinct scenarios. Each reported value is averaged over $4\times100 = 400$ trials, with the variation across seeds also provided to indicate the stability of policy learning. The results are shown in Fig.~\ref{fig:sim_sr} and Fig.~\ref{fig:sim_exp}.

From the figures, we observe that our method continuously improves the policy over multiple rounds of self-improvement, achieving substantial gains in success rate with relatively few rollouts. Notably, it is the only approach that demonstrates consistent improvement across all tasks. In contrast, both baselines exhibit unstable performance, with success rates fluctuating or even declining in certain tasks. Regarding sample efficiency, although our method requires a similar number of rollouts as the baselines in the first round, it quickly reduces interaction costs in subsequent rounds, indicating that the policy effectively leverages each interaction and avoids wasting samples on uninformative exploration behaviors. Overall, these results demonstrate that our exploration mechanism can robustly improve policy performance across diverse tasks and over multiple iterations, highlighting its potential for scalable robot learning.

\subsection{Ablation Study}

\begin{figure}[t]
    \centering
    \includegraphics[width=0.48\linewidth]{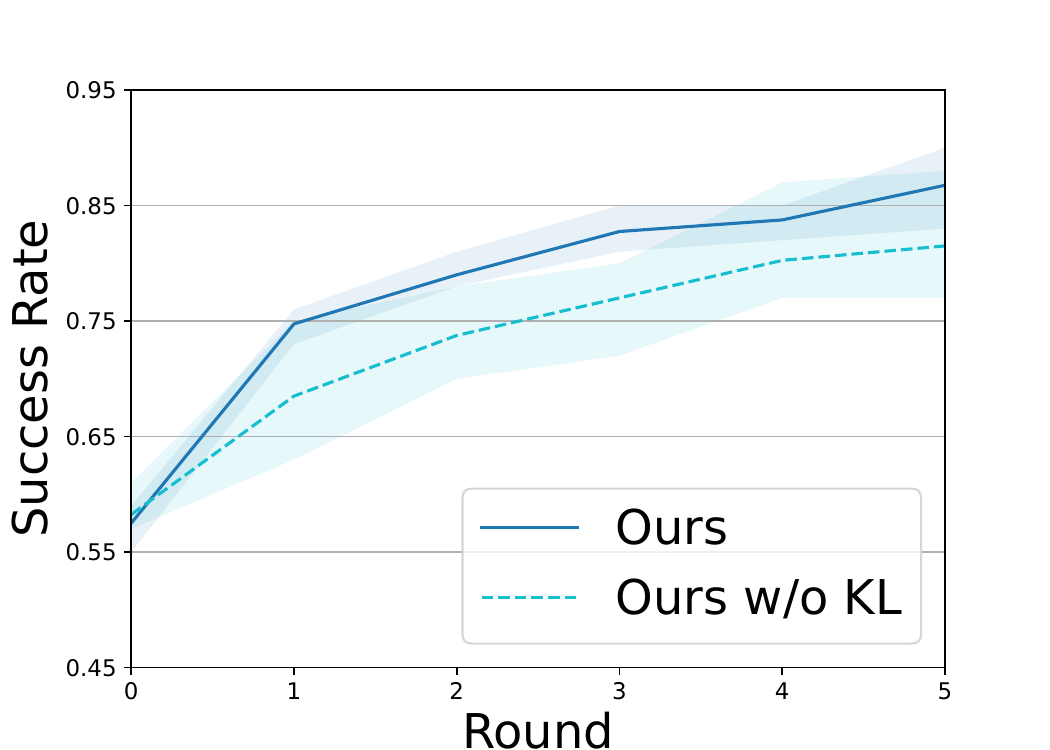}
    \includegraphics[width=0.48\linewidth]{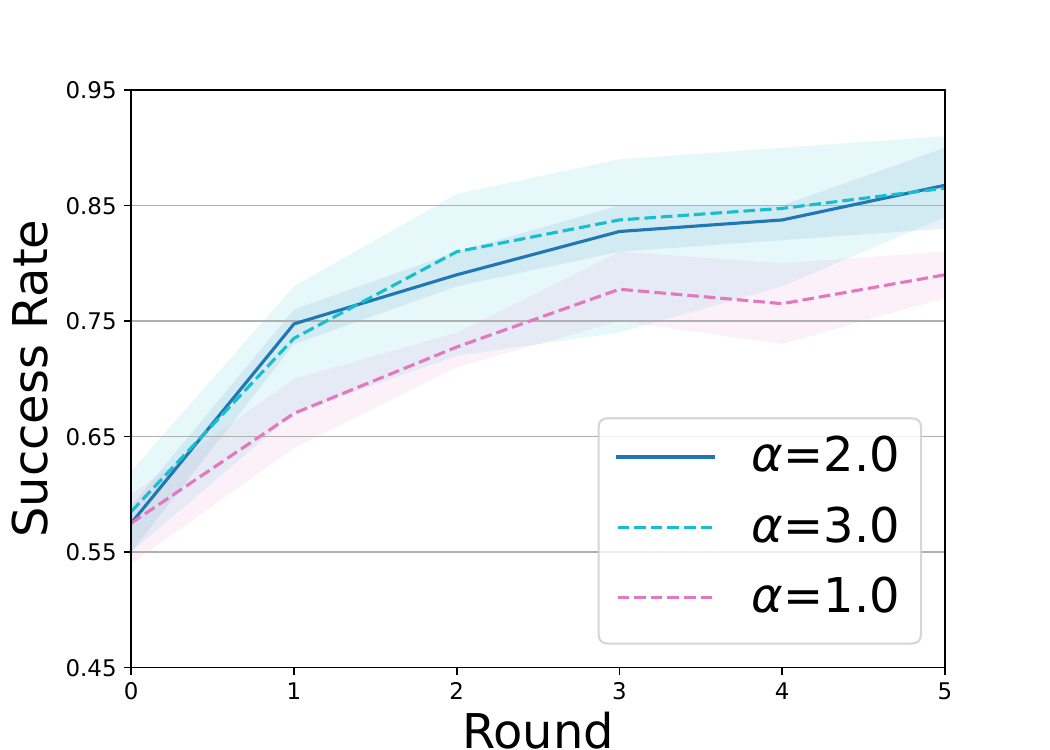}
    \includegraphics[width=0.48\linewidth]{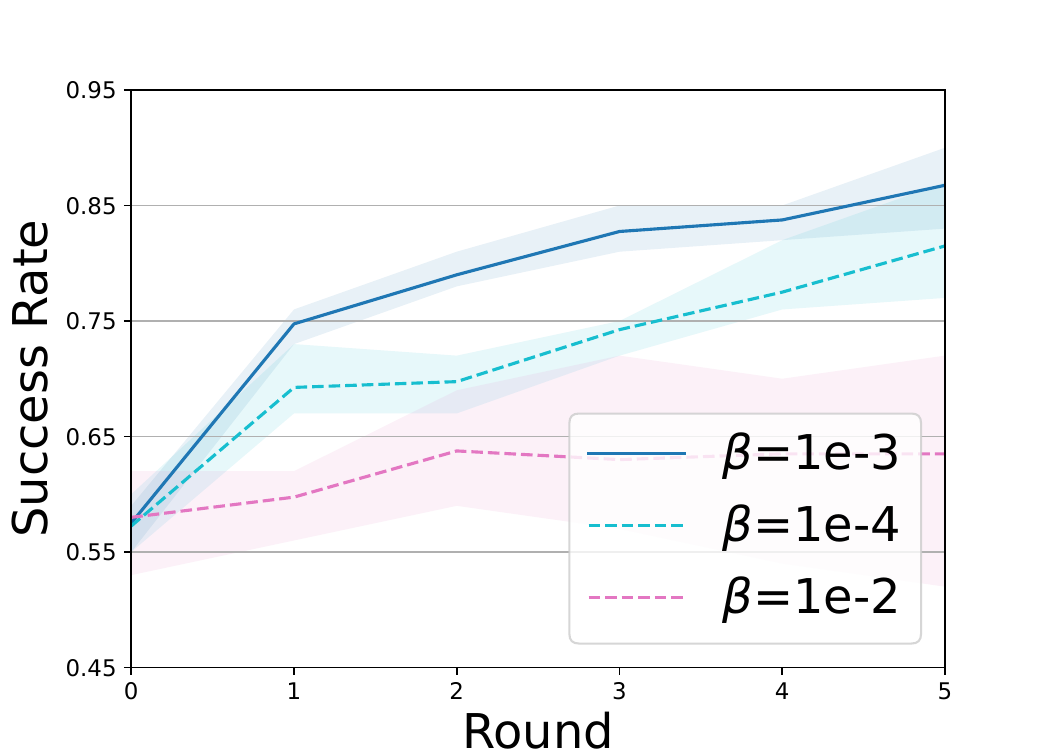}
    \includegraphics[width=0.48\linewidth]{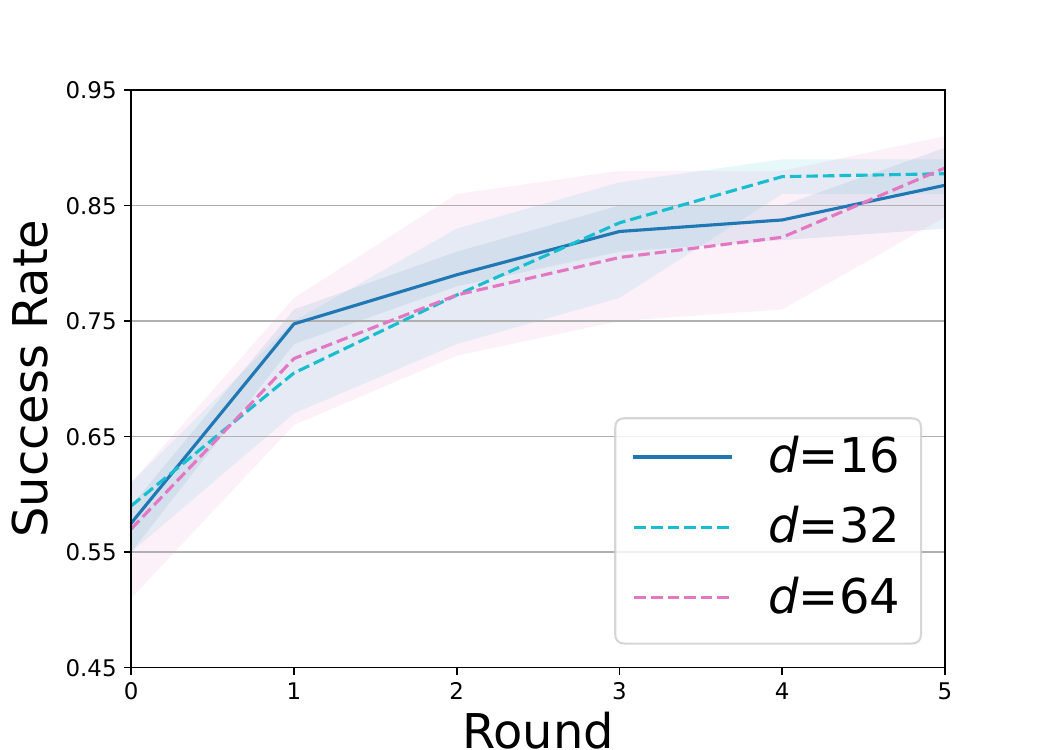}
    \caption{\textbf{Task success rate under ablation.} We conduct these ablations on the \textit{Can-20} task.}
    \vspace{-5mm}
    \label{fig:ablation}
\end{figure}

To better understand the contributions of different components and the sensitivity to hyperparameters, we conduct an ablation study on the \textit{Can} task from the RoboMimic benchmark. The results are presented in Fig.~\ref{fig:ablation}.

We begin by examining the impact of the KL term in $\mathcal{L}_{\text{IB}}$. As shown in the upper-left plot, removing the KL term reduces the final success rate from 86.75\% to 81.50\%, highlighting the importance of enforcing a compact and informative latent representation for effective exploration.

We then analyze the effects of the noise scale $\alpha$ and the KL weight $\beta$. The upper-right and bottom-left plots show that both hyperparameters play a critical role in balancing exploration diversity and stability. A small noise scale $\alpha$ leads to conservative exploration and sluggish improvement, whereas an excessively large $\alpha$ induces aggressive strategies and unstable improvement. 
Likewise, setting the KL weight $\beta$ too low results in entangled latent representations and insufficient diversity, while too high a $\beta$ can cause model collapse, failing to generate meaningful actions. To achieve a trade-off between improvement magnitude and stability, we adopt $\alpha=2.0$ and $\beta=0.001$ for most simulation tasks.

We further evaluate the robustness of our method to the latent dimension $d$. As shown in the bottom-right plot, varying $d$ from 16 to 64 has little effect on the final success rate, suggesting that the VIB objective effectively constrains the latent space to capture only task-relevant information, independent of its nominal dimensionality. Interestingly, the number of effective dimensions, measured by SNR, remains invariant across different $d$. For example, in the \textit{Can} task, the effective dimension consistently stays at 8 even when $d$ ranges from 16 to 64. Moreover, the effective dimension appears to correlate with task complexity: simple tasks such as \textit{Lift} require few dimensions, while more complex tasks such as \textit{Transport} require more, as summarized in Table~\ref{tab:dim}.

\begin{table}[htbp]
    \centering
    \caption{\textbf{Number of Effective Dimensions.}}
    \begin{tabular}{ccccc}
        \toprule
        & \textit{Lift} & \textit{Can} & \textit{Square} & \textit{Transport} \\
        \midrule
        \textbf{Intrinsic Dim} & 8 & 8 & 10 & 16 \\
        \bottomrule
    \end{tabular}
    \label{tab:dim}
\end{table}

Taken together, these results suggest the existence of a policy-agnostic intrinsic dimension for each task, which can be interpreted as the minimal degrees of freedom required to transfer task-relevant information from observations to actions. Our method is able to automatically identify and exploit these intrinsic dimensions, enabling structured on-manifold exploration that enhances both the efficiency and effectiveness of policy self-improvement.



\section{Conclusion}

In this paper, we present \textit{SOE}, a novel exploration approach for sample-efficient robot policy self-improvement. By integrating a variational information bottleneck with policy learning as an auxiliary plug-in module, our method learns a compact latent representation that enables structured on-manifold exploration, while preserving the integrity of the base policy performance. Our experiments demonstrate that this approach leads to more effective, safer, and more efficient exploration compared to prior methods. Furthermore, the learned latent space facilitates user-guided steering, allowing users to intuitively direct exploration toward desired behaviors. Overall, our work highlights the potential of structured exploration mechanisms, hoping to inspire future research in the development of sample-efficient and self-improving robot learning systems.








\printbibliography

\end{document}